\documentclass[letterpaper,10pt,conference]{ieeeconf}

\IEEEoverridecommandlockouts
\usepackage{etoolbox} 
\usepackage[utf8]{inputenc} 
\usepackage[T1]{fontenc}    
\usepackage[english]{babel} 
\usepackage{comment}
\usepackage{cancel}
\usepackage{csquotes}
\usepackage{listings}
\usepackage{nameref}
\usepackage{paralist}
\usepackage{xspace}
\usepackage{setspace}
\usepackage{makeidx}
\usepackage{pdfpages}
\usepackage{upgreek}
\usepackage{balance}

\usepackage{amsmath,amssymb} 
\usepackage{amsfonts}       
\usepackage{mathtools}
\usepackage{array} 
\usepackage{nicefrac}       
\usepackage{breqn}          
\usepackage{textcomp}
\usepackage{bm} 
\usepackage{pifont}

\usepackage{graphicx}
\usepackage{epsfig}

\usepackage{float}
\usepackage{subcaption}
\usepackage[font=small,labelfont=bf]{caption}
\usepackage{lscape}      

\usepackage{tikz}  
\usepackage{makecell}
\usepackage{algorithm}

\usepackage{array} 
\usepackage{tabularx}
\usepackage{multirow}
\usepackage{multicol}
\usepackage{booktabs}   

\usepackage{paralist}
\let\labelindent\relax 
\usepackage{enumitem}
\setlist[itemize]{noitemsep,leftmargin=*,topsep=0in}
\setlist[enumerate]{noitemsep,leftmargin=*,topsep=0in}

\usepackage{url}            
\usepackage{color,colortbl} 
\usepackage{soul}

\PassOptionsToPackage{capitalize, noabbrev}{cleveref}
\usepackage{hyperref}
\hypersetup{pagebackref=false,breaklinks=true,colorlinks,urlcolor=blue,citecolor=blue,linkcolor=blue,bookmarks=false}

\makeatletter
\let\NAT@parse\undefined
\makeatother
\usepackage[numbers,sort&compress]{natbib}

\definecolor{selfbg}{RGB}{211,228,204}
\definecolor{pursuersbg}{RGB}{227,202,203}
\definecolor{evadersbg}{RGB}{200,200,224}
\definecolor{obstaclesbg}{RGB}{187,189,192}

\usepackage{amsthm}

\usepackage{graphicx}
\usepackage{amsmath,amssymb}
\usepackage{booktabs}
\usepackage{cite}

\title{\LARGE \bf
Higher-Order Morphology Priors for Quadruped Reinforcement Learning Under Actuator Degradation}
\author{Derek You$^{*1}$, Zafir Shamsi$^{*2}$, Keqin Wang$^{*\dagger3}$ and Christine Allen-Blanchette$^{\dagger3}$
\thanks{$^*$Equal contribution. $^{1}$Edison Academy Magnet School. $^{2}$South Brunswick High School. Work done during the Princeton Laboratory Learning Program. $^{3}$Princeton University. $^{\dagger}$Correspondence: {\tt\small \{keqin.wang, ca15\}@princeton.edu}}
}

\begin{document}

\maketitle
\thispagestyle{empty}
\pagestyle{empty}

\begin{abstract}
Actuator degradation turns quadruped locomotion into a coordination problem requiring joints to compensate for lost actuation. Prior work suggests that morphology-aware graph policies improve learning and generalization under body perturbations. We ask whether these benefits can be strengthened by explicitly modeling higher-order mechanical structure. We represent the Unitree Go1 as a cell complex with limb- and body-level rank-2 cells and apply Hodge-based message passing. Under degradation training, the node-edge-face Hodge actor achieves the highest return on unseen actuator degradations, with higher survival and lower velocity-tracking error. These results support higher-order morphology as a useful inductive bias for whole-body compensation under actuator degradation.

\end{abstract}

\section{Introduction}

Learning-based controllers have made quadruped locomotion increasingly robust to terrain variation, uncertain dynamics, and changes in external loading \citep{hwangbo2019agile,kumar2021rma}. Actuator degradation instead directly reduces the control authority available at specific joints, requiring the remaining body to compensate. Existing fault-tolerant approaches address this challenge through explicit mechanisms for adapting to the fault \citep{cully2015adapt,hou2024fault,dreamflex2025,xu2025acl,gravina2026fault}. This motivates the question: rather than adding dedicated fault-handling mechanisms, can robustness emerge simply from how the quadruped is represented in the policy?

Morphology-aware policies show that representations of robot morphology shape locomotion and transfer \citep{wang2018nervenet,huang2020smp,chiappa2022dmap,sferrazza2025bot}. Most such work encodes morphology as a classical graph, with joints as nodes and links as edges. At a higher level, joints form mechanical units such as limbs and body-level structures. Such a graph can represent the edges bounding these structures, but not the structures themselves as higher-order objects. We therefore ask whether enriching the graph with mechanically meaningful higher-order structure provides a useful inductive bias for coordinated compensation under actuator degradation.

To investigate this question, we represent the Unitree Go1 as a cell complex whose rank-2 cells encode four limb-level structures and one body-level structure, and apply Hodge-based message passing over nodes, edges, and faces. The resulting faces make these multi-joint units explicit rather than leaving them implicit in their bounding edges.

The main contributions of this work are twofold. (1) We introduce a higher-order morphology representation for the Unitree Go1 with mechanically motivated rank-2 limb and body cells, and a node-edge-face Hodge actor over it. (2) Through a capacity-matched healthy/degradation training-evaluation study, we show that our approach outperforms competitive baselines on held-out compound actuator degradation data.

\begin{figure}[t]
    \centering
    \includegraphics[width=0.75\columnwidth]{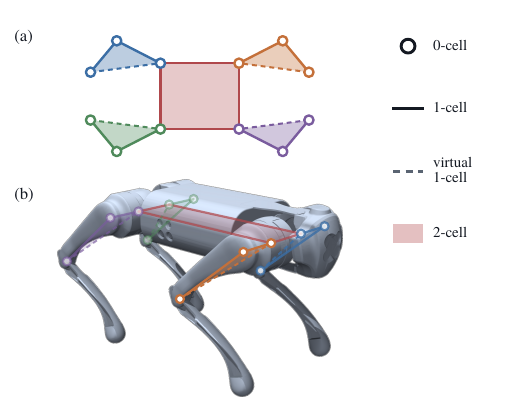}
    \caption{\textbf{Go1 higher-order morphology.} \textbf{(a)} The twelve actuated joints form the 0-cells; sixteen 1-cells give eight serial
couplings, four around the hip ring, and four dashed virtual closures that let each leg bound a
face. The five shaded 2-cells are one per limb plus the body cell spanned by the four coplanar
hips. \textbf{(b)} The same cells at the true Go1 joint positions, with distant limbs
faded. Only the node-edge-face actor propagates through rank-2 cells.
The legend maps each mark to its rank: a ring is a 0-cell, a solid line a 1-cell, a dashed line
a virtual 1-cell, and a filled patch a 2-cell.
}
    \label{fig:cell_complex}
    \vspace{-15pt}
\end{figure}

\section{Related Work}

\textbf{Learned quadruped locomotion.}
Reinforcement learning now yields quadruped controllers that traverse demanding terrain
\citep{lee2020learning,miki2022learning}, transfer to hardware
\citep{tan2018sim,hwangbo2019agile}, adapt online \citep{kumar2021rma}, and train in minutes
\citep{rudin2022learning,margolis2023walk}. These are typically centralized networks over a flat
observation vector, leaving the body's mechanical organization implicit.

\textbf{Fault-tolerant locomotion.}
Robot damage adaptation has been studied through both online search and learning-based control. Cully et al.~\citep{cully2015adapt} rapidly adapt to physical damage by searching a repertoire of precomputed behaviors. Recent quadruped methods instead incorporate faults into policy learning, including hierarchical fault-tolerant control \citep{hou2024fault}, fault-conditioned locomotion \citep{dreamflex2025}, teacher-student distillation across compensatory gaits \citep{xu2025acl}, and gait adaptation under actuator power loss \citep{gravina2026fault}. Rather than introducing a new adaptation mechanism, we ask whether the policy's morphology representation itself facilitates coordinated compensation under a fixed learning protocol.

\textbf{Morphology-aware locomotion.}
Morphology-aware policies organize computation according to an agent's articulated body, typically using graph-based architectures \citep{gilmer2017neural,battaglia2018relational,bronstein2021geometric}. NerveNet~\citep{wang2018nervenet} and Shared Modular Policies~\citep{huang2020smp} showed that morphology-structured communication improves transfer across embodiments and disabilities, and DMAP \citep{chiappa2022dmap} found these benefits become more pronounced under larger body perturbations. Body Transformer \citep{sferrazza2025bot}, however, showed that purely local morphological communication can bottleneck information propagation on harder tasks. These works establish morphology as a useful prior, but model it through pairwise relations. We instead enrich it with mechanically meaningful higher-order relations.

\textbf{Higher-order graph learning.}
Higher-order neural architectures extend graph learning beyond pairwise edges by placing signals on edges, faces, and higher-dimensional cells and coupling adjacent ranks through incidence and Hodge operators \citep{ebli2020simplicial,bodnar2021mpsn,bodnar2021cw,yang2022simplicial}. Simplicial neural networks and message-passing simplicial networks generalize graph propagation to simplicial complexes, while cellular and Hodge-based constructions further support communication among cells of different ranks \citep{hajij2023topological,papillon2023architectures}; the underlying signal processing rests on Hodge Laplacians \citep{lim2020hodge,barbarossa2020topological,schaub2020random}. We use this machinery in a different role: rank-2 cells represent persistent mechanical structures of a quadruped, and higher-order message passing serves as a policy-level morphology prior whose effect is evaluated under actuator degradation.

\section{Preliminaries}
\label{sec:preliminaries}

\subsection{Unitree Go1}
\label{sec:go1}
The Unitree Go1 is a 12-DoF quadruped with four three-joint legs attached to a rigid trunk (see Fig.~\ref{fig:cell_complex}). Each leg comprises a hip abduction/adduction joint and hip and knee flexion/extension joints.

Two properties motivate the higher-order representation in Sec.~\ref{sec:cell_complexes}. First, each leg acts as a mechanical unit: foot placement depends jointly on all three joints, so lost actuation at one must be compensated for by the others. Second, the four hips are rigidly mounted in a fixed coplanar arrangement, forming a configuration-independent hip ring with a shared reference frame.

\subsection{Problem Setting}
\label{sec:problem_setting}

We consider velocity-commanded locomotion using the MuJoCo Playground \texttt{Go1JoystickFlatTerrain} task \citep{mujoco_playground_2025}. At each control step $t$, the actor receives
an observation $o_t\in\mathbb{R}^{48}$ containing base quantities (linear and
angular velocity and projected gravity), the commanded velocity, and per-joint
quantities (positions, velocities, and previous actions), and outputs
$a_t\in\mathbb{R}^{12}$, comprising one normalized position offset for each
actuator. Policies are trained with proximal policy optimization (PPO)
\citep{schulman2017ppo} using the task's provided locomotion reward.

We model degradation of actuator $i$ by scaling its maximum available torque
$\tau_i^{\mathrm{max}}$,
\[
    \tau_i^{\mathrm{max}}(\rho_i)
    =
    \rho_i \tau_i^{\mathrm{max,nom}},
    \qquad
    \rho_i\in[0,1],
\]
where $\tau_i^{\mathrm{max,nom}}$ is the nominal torque limit of actuator $i$,
$\rho_i=1$ denotes nominal actuation, and $\rho_i=0$ denotes complete loss of available torque.

\subsection{Cell Complexes and Incidence Operators}
\label{sec:cell_complexes}

A classical graph consists of nodes and edges, so its structure is fully
described by rank-0 and rank-1 cells. A cell complex retains this node-edge
structure but additionally promotes selected closed edge cycles to explicit
rank-2 cells, which we refer to as faces
\citep{bodnar2021mpsn,yang2022simplicial}. 

Let $\mathcal{X}_0$,
$\mathcal{X}_1$, and $\mathcal{X}_2$ denote the sets of nodes, edges, and
faces, respectively.
Signals defined on rank-$k$ cells are $k$-cochains, represented in our
networks by feature matrices $H_k$ associated with nodes, edges, or faces.
Connectivity between adjacent ranks is encoded by the oriented incidence
matrices
\[
    B_1 \in
    \mathbb{R}^{|\mathcal{X}_0|\times|\mathcal{X}_1|},
    \qquad
    B_2 \in
    \mathbb{R}^{|\mathcal{X}_1|\times|\mathcal{X}_2|}.
\]
These operators and their transposes mediate message passing, cell-complex message passing updates features within each rank while exchanging information across ranks through adjacent-rank incidence operators, which can be summarized as
\[
    H_0
    \;\overset{B_1^\top}{\underset{B_1}{\rightleftarrows}}\;
    H_1
    \;\overset{B_2^\top}{\underset{B_2}{\rightleftarrows}}\;
    H_2.
\]
\begin{figure*}[t]
  \centering
  \includegraphics[width=0.9\textwidth]{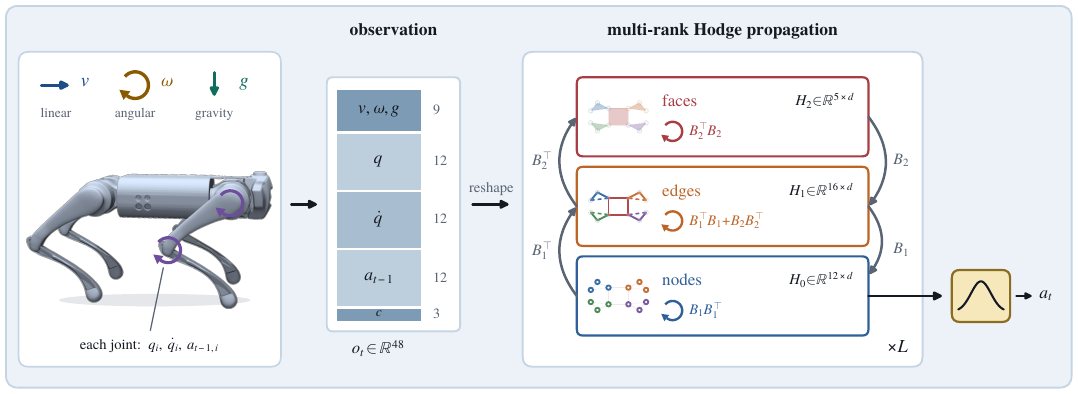}
\caption{\textbf{Node-edge-face Hodge actor.}
\textbf{Left:} the policy observes trunk velocity, angular velocity, projected gravity, and joint-wise angle, velocity, and previous action.
\textbf{Centre:} these signals and the velocity command form $o_t\in\mathbb{R}^{48}$ and are reshaped into joint-wise features $X_t\in\mathbb{R}^{12\times20}$.
\textbf{Right:} $X_t$ initializes node, edge, and face states
$H_0\in\mathbb{R}^{12\times d}$, $H_1\in\mathbb{R}^{16\times d}$, and
$H_2\in\mathbb{R}^{5\times d}$. Each Hodge block performs same-rank message passing through the corresponding Hodge operators and cross-rank communication through incidence matrices $B_1$ and $B_2$. After $L$ residual blocks, the rank-0 states are decoded by a shared actor head to parameterize the joint action $a_t\in\mathbb{R}^{12}$.}
  \label{fig:hodge_actor}
  \vspace{-10pt}
\end{figure*}

\section{Method}
\label{sec:method}

\subsection{Higher-Order Go1 Representation}
\label{sec:go1_complex}

The Go1 cell-complex construction is shown in
Fig.~\ref{fig:cell_complex}. Each node (0-cell) corresponds to one of the twelve joints
on the robot. Eight kinematic edges (1-cell) connect adjacent joints along the four
legs. Four additional edges connect the hip joints around the body, while
four hip-calf edges close the four limb cycles. These cycles define four limb faces (2-cell) and one body face (2-cell), yielding a cell
complex with 12 nodes, 16 edges, 5 faces, and fixed incidence matrices
$B_1\in\mathbb{R}^{12\times16}$ and
$B_2\in\mathbb{R}^{16\times5}$.

\subsection{Feature Initialization}
\label{sec:features}

Rather than processing the actor observation $o_t$ as a single vector, we
reorganize it into a joint feature matrix
\[
    X_t
    =
    \begin{bmatrix}
        x_{t,1},\dots,x_{t,12}
    \end{bmatrix}^{\top}
    \in\mathbb{R}^{12\times20},
\]
where each joint feature $x_{t,i}\in\mathbb{R}^{20}$ contains the shared base
quantities and velocity command from $o_t$, the position, velocity, and
previous action corresponding to joint $i$, and additional static indicators
of limb and joint type.

Each feature is projected into the hidden dimension to initialize the node
features $H_0^{(0)}$. Edge features $H_1^{(0)}$ come from the embeddings of their incident joints plus static edge information.
For the node-edge-face actor, face features $H_2^{(0)}$ additionally use the oriented boundary aggregate $B_2^\top H_1^{(0)}$
plus static face information.

\subsection{Multi-Rank Hodge Message Passing}
\label{sec:hodge_message_passing}

Each propagation layer combines \emph{same-rank} messages $S_k$ with
\emph{cross-rank} messages $C_k$. Same-rank propagation is defined by the
Hodge Laplacian
\[
\begin{aligned}
    L_k
    &=
    \underbrace{B_k^\top B_k}_{L_k^{\mathrm{lower}}}
    +
    \underbrace{B_{k+1}B_{k+1}^\top}_{L_k^{\mathrm{upper}}},
    \qquad
    S_k
    =
    L_k H_k.
\end{aligned}
\]
The lower component maps a rank-$k$ signal to its incident lower-rank cells
and back, allowing rank-$k$ cells that share a boundary to exchange
information. The upper component similarly propagates through incident
higher-rank cells and back, coupling rank-$k$ cells that share a coface.

Cross-rank messages take the form
\[
    C_k
    =
    \underbrace{B_k^\top H_{k-1}}_{\text{from }k-1}
    +
    \underbrace{B_{k+1} H_{k+1}}_{\text{from }k+1},
\]
and directly aggregate information from adjacent ranks into rank $k$.
Terms involving nonexistent adjacent ranks are omitted; e.g.,
$B_0^\top$ or $H_{-1}$.

At each layer $\ell$ and rank $k$, a learned mixing coefficient
$\alpha_k^{(\ell)}\in(0,1)$ balances the two message types,
\[
    M_k^{(\ell)}
    =
    \alpha_k^{(\ell)}S_k^{(\ell)}
    +
    \bigl(1-\alpha_k^{(\ell)}\bigr)C_k^{(\ell)}.
\]
The mixed message is passed through a learned norm gate and root-mean-square
layer normalization (RMSNorm) \citep{zhang2019rmsnorm} before a residual
update,
\[
    H_k^{(\ell+1)}
    =
    H_k^{(\ell)}
    +
    \operatorname{RMSNorm}\!\left(
        g_k^{(\ell)}(M_k^{(\ell)})
    \right).
\]

\subsection{Policy and Value Networks}
\label{sec:policy_decoding}

After four propagation layers, the same two-layer decoder
$f_{\mathrm{dec}}$ is applied independently to each final joint embedding
$\left[H_0^{(L)}\right]_{i,:}$ to obtain the parameters of its action
distribution,
\[
    (\mu_{t,i},\sigma_{t,i})
    =
    f_{\mathrm{dec}}\!\left(
        \left[H_0^{(L)}\right]_{i,:}
    \right),
\]
where $\mu_{t,i}$ and $\sigma_{t,i}>0$ denote the mean and standard deviation
for joint $i$. The resulting tanh-transformed Gaussian distributions define
the factorized policy
\[
    \pi_\theta(a_t\mid o_t)
    =
    \prod_{i=1}^{12}
    \operatorname{TanhNormal}
    \!\left(a_{t,i};\mu_{t,i},\sigma_{t,i}\right),
\]
where $\theta$ denotes the actor parameters.

The critic receives a privileged observation
$o_t^{\mathrm{priv}}\in\mathbb{R}^{123}$ containing $o_t$ along with additional body, joint, actuator, and contact state available from the simulator, and estimates the state value as
\[
    V_\phi(o_t^{\mathrm{priv}})
    =
    \operatorname{MLP}_{\phi}(o_t^{\mathrm{priv}}),
\]
where $\phi$ denotes the critic parameters. We use this standard MuJoCo
Playground critic for all actor variants to isolate differences among the actor architectures.

\section{Experiments}
\subsection{Experimental Setup and Baselines}

All experiments use the Unitree Go1 problem setting described in Sec.~\ref{sec:problem_setting}. All actor variants share the same critic, reward, domain randomization, optimization setup, and training budget. Each policy is trained for 400M environment steps with seeds $\{11,12,13\}$.

We compare five parameter-matched actor architectures:
\begin{itemize}\setlength{\itemsep}{1pt}
    \item \textbf{MLP-C}: centralized MLP mapping the global 48-D observation directly to all joint outputs.
    \item \textbf{MLP-D}: factorized per-joint MLP with shared weights and no inter-joint message passing.
    \item \textbf{GCN}: two-layer graph convolutional network \citep{kipf2017gcn}.
    \item \textbf{Hodge-L}: node-edge Hodge actor introducing multi-rank propagation over ranks 0-1.
    \item \textbf{Hodge-F}: node-edge-face Hodge actor extending Hodge-L with rank-2 cells.
\end{itemize}
MLP-D, GCN, Hodge-L, and Hodge-F use the same 20-D per-joint inputs described in Sec.~\ref{sec:features}.

\begin{figure*}[!t]
    \centering
    \begin{minipage}[t]{0.85\columnwidth}
        \centering
        \includegraphics[width=\linewidth]{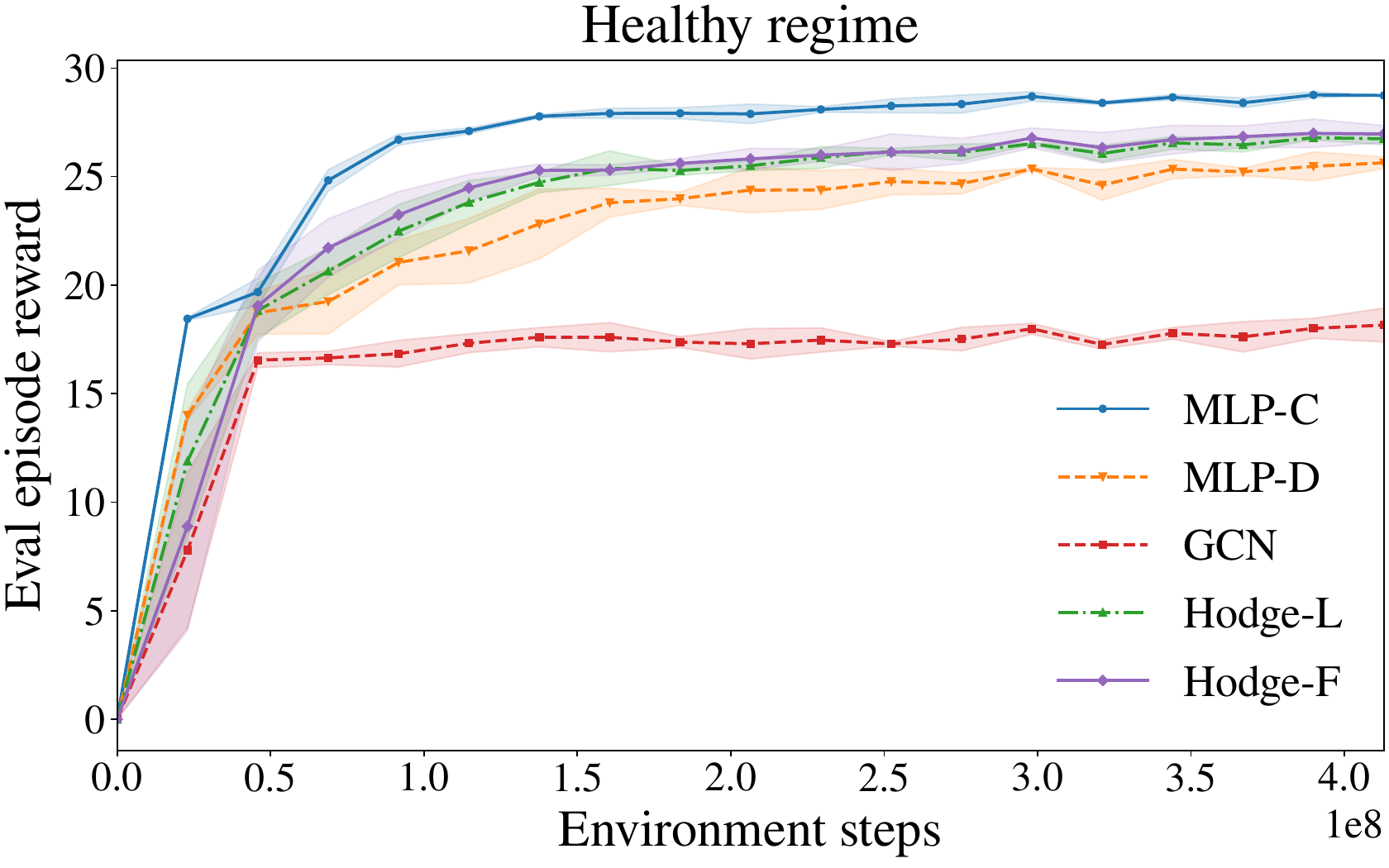}
    \end{minipage}
    \hfill
    \begin{minipage}[t]{0.85\columnwidth}
        \centering
        \includegraphics[width=\linewidth]{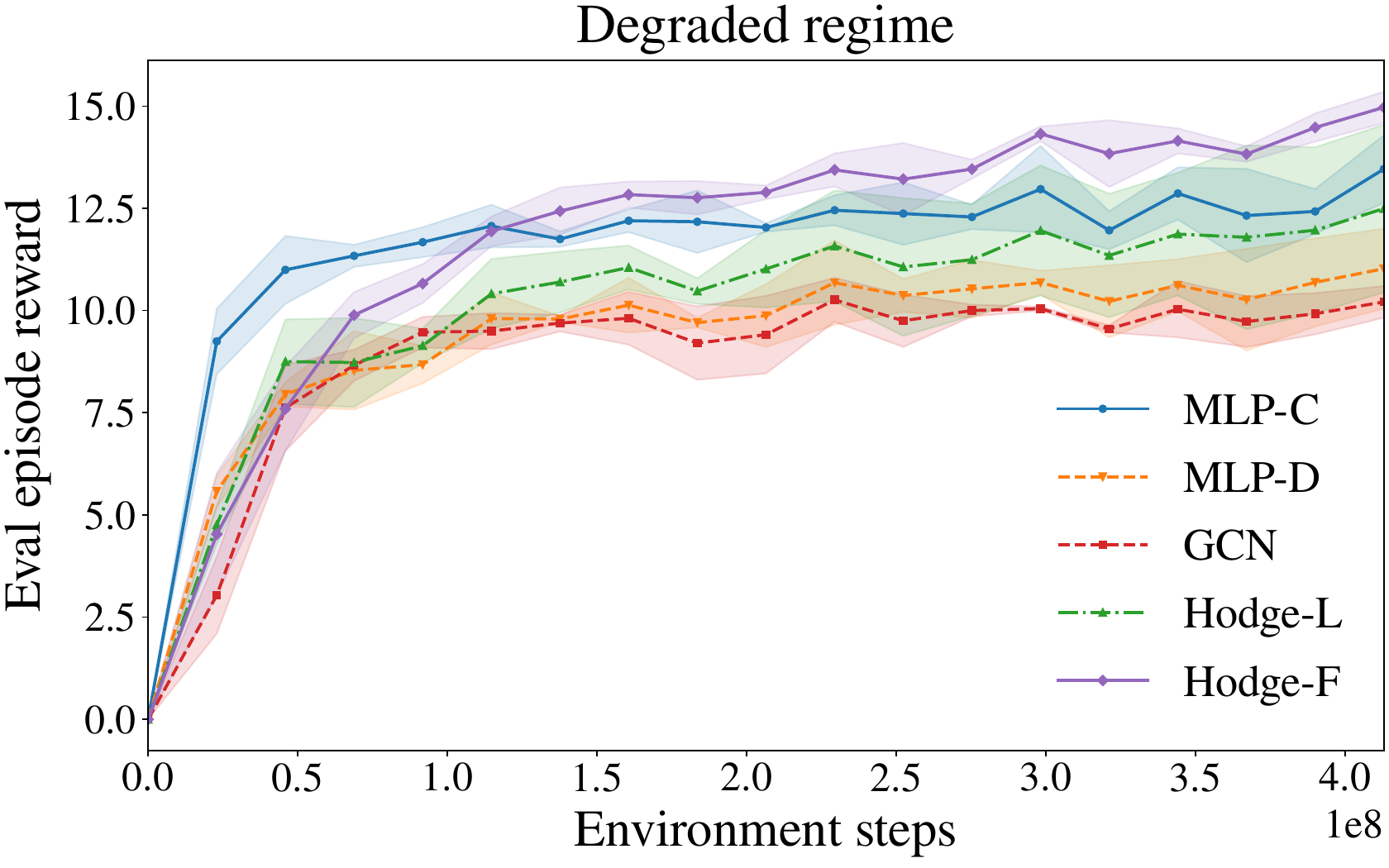}
    \end{minipage}
    \caption{\textbf{In-regime learning curves.} Evaluation episode returns under the healthy (left) and degraded (right) regimes. In the healthy regime, MLP-C performs best, with Hodge-F and Hodge-L slightly behind. Under degradation, Hodge-F achieves the strongest performance.}

    \label{fig:learning_curves}
\end{figure*}

\subsection{Training and Evaluation Regimes}
We design two experimental regimes to compare our node-edge-face method against the baselines:
\begin{itemize}\setlength{\itemsep}{1pt}
    \item \textbf{Healthy regime ($H$):} No actuator degradation: $\rho_i=1$ for all actuators during training and evaluation.    
    \item \textbf{Degraded regime ($D$):} We degrade two actuators at a time, giving $\binom{12}{2}=66$ unordered pairs, split into 44 training and 22 held-out pairs. During training, the pair is fixed per episode and sampled as $30\%$ healthy ($\rho_i=\rho_j=1$), $50\%$ weak ($\rho_i,\rho_j\sim\operatorname{LogUniform}(0.05,0.20)$), or $20\%$ dead ($\rho_i=\rho_j=0$). Evaluation averages over all held-out pairs and all $\rho_i,\rho_j\in\{0,0.025,0.05,0.075,0.10,0.15,0.20,0.25,0.50\}$.
\end{itemize}

To evaluate the performance and generalization of our method, we conduct 4 experiments: in-regime training and evaluation, denoted as $H\to H$ and $D\to D$, and cross-regime evaluation, denoted as $H\to D$ and $D\to H$.

\subsection{In-Regime Training and Evaluation}

We first compare the matched training-evaluation settings, $H\!\to\!H$ and $D\!\to\!D$. Fig.~\ref{fig:learning_curves} shows the corresponding learning curves. MLP-C improves rapidly early in training under both curricula and retains its lead under healthy regime training. Under the degraded regime, however, Hodge-F eventually surpasses the other methods and achieves the highest episode return, indicating that the morphology representation prior improves performance in this setting.


\begin{table}[!ht]
\centering
\caption{\textbf{In-regime evaluation results.} Cumulative reward, survival rate, and velocity error are reported for $H\to H$ (left) and $D\to D$ (right) across different actors.}
\label{tab:healthy_metrics}
\scriptsize
\resizebox{\columnwidth}{!}{%
\begin{tabular}{l|ccc|ccc}
\hline
\multirow{2}{*}{\textbf{Actor}}
& \multicolumn{3}{c|}{$H \to H$}
& \multicolumn{3}{c}{$D \to D$} \\
\cline{2-7}
& Return $\uparrow$ & Survival $\uparrow$ & Vel. RMSE $\downarrow$ & Return $\uparrow$ & Survival $\uparrow$ & Vel. RMSE $\downarrow$\\
\hline
MLP-C   & \textbf{33.88$\pm$0.03} & 1.00$\pm$0.00 & \textbf{0.052$\pm$0.001} & 19.98$\pm$0.50 & 0.88$\pm$0.02 & 0.324$\pm$0.003\\
MLP-D   & 31.17$\pm$0.15 & 1.00$\pm$0.00 & 0.060$\pm$0.001 & 18.15$\pm$1.09 & 0.83$\pm$0.06 & 0.320$\pm$0.008\\
GCN     & 24.52$\pm$0.07 & 1.00$\pm$0.00 & 0.302$\pm$0.001 & 15.35$\pm$0.21 & 0.73$\pm$0.01 & 0.339$\pm$0.001\\
Hodge-L & 32.30$\pm$0.27 & 1.00$\pm$0.00 & 0.057$\pm$0.002 & 18.61$\pm$1.79 & 0.83$\pm$0.09 & 0.325$\pm$0.008\\
Hodge-F & 32.58$\pm$0.53 & 1.00$\pm$0.00 & 0.056$\pm$0.002 & \textbf{22.40$\pm$0.63} & \textbf{0.94$\pm$0.01} & \textbf{0.252$\pm$0.036}\\
\hline
\end{tabular}
}
\end{table}

The in-regime evaluation results are reported in Table~\ref{tab:healthy_metrics}. Return is the cumulative locomotion reward over an evaluation rollout. Survival rate is the proportion of episodes in which the robot does not fall over. Velocity RMSE is the root-mean-square error between commanded and actual velocity. In $H\to H$, MLP-C achieves the highest return and lowest velocity error. In $D\to D$, Hodge-F achieves the highest return, survival rate, and lowest velocity error, demonstrating robustness under degradation.

MLP-D and GCN ablate the sources of Hodge-F's gain over MLP-C under $D\to D$. Compared with Hodge-F, MLP-D achieves a lower return and a higher velocity error, indicating that factorization alone does not explain the gain. GCN likewise yields a lower return and a higher velocity error, highlighting the benefit of the Hodge-based representation beyond a conventional graph.

Hodge-F also improves survival and reduces velocity RMSE relative to Hodge-L under $D\to D$ setting. Thus, the advantage of faces emerges primarily when learned compensation is required, rather than under nominal locomotion. This is also reflected in Fig.~\ref{fig:learning_curves}, where Hodge-F clearly outperforms Hodge-L only in the degraded regime.

\subsection{Cross-Regime Evaluation}

To evaluate generalization, we swap the evaluation regimes without retraining, yielding $H\!\to\!D$ and $D\!\to\!H$. Table~\ref{tab:cross_regime_metrics} summarizes the performance.


\begin{table}[!ht]
\centering
\caption{\textbf{Cross-regime evaluation results.} Cumulative reward, survival rate, and velocity error are reported for $H\to D$ (left) and $D\to H$ (right) across different actors.}
\label{tab:cross_regime_metrics}
\scriptsize
\resizebox{\columnwidth}{!}{%
\begin{tabular}{l|ccc|ccc}
\hline
\multirow{2}{*}{\textbf{Actor}}
& \multicolumn{3}{c|}{$H \to D$}
& \multicolumn{3}{c}{$D \to H$} \\
\cline{2-7}
& Return $\uparrow$
& Survival $\uparrow$
& Vel. RMSE $\downarrow$
& Return $\uparrow$
& Survival $\uparrow$
& Vel. RMSE $\downarrow$ \\
\hline
MLP-C
& \textbf{16.89$\pm$0.10}
& 0.69$\pm$0.05
& 0.205$\pm$0.005
& 24.85$\pm$0.21
& 1.00$\pm$0.00
& 0.306$\pm$0.001 \\

MLP-D
& 13.87$\pm$0.41
& 0.57$\pm$0.03
& 0.208$\pm$0.003
& 24.74$\pm$0.32
& 1.00$\pm$0.00
& 0.307$\pm$0.004 \\

GCN
& 12.99$\pm$0.28
& \textbf{0.70$\pm$0.02}
& 0.342$\pm$0.009
& 23.83$\pm$0.18
& 1.00$\pm$0.00
& 0.307$\pm$0.001 \\

Hodge-L
& 15.32$\pm$0.92
& 0.63$\pm$0.03
& \textbf{0.195$\pm$0.017}
& 24.87$\pm$0.31
& 1.00$\pm$0.00
& 0.311$\pm$0.000 \\

Hodge-F
& 15.59$\pm$1.16
& 0.60$\pm$0.03
& 0.207$\pm$0.001
& \textbf{27.17$\pm$1.73}
& 1.00$\pm$0.00
& \textbf{0.220$\pm$0.060} \\
\hline
\end{tabular}
}
\end{table}

In $H\!\to\!D$, MLP-C achieves the highest return, outperforming Hodge-F under zero-shot degradation. This indicates that morphology-aware structure alone is insufficient to guarantee robustness to unseen actuator failures when the policy is trained only under nominal conditions. In $D\!\to\!H$, however, Hodge-F achieves both the highest return and lowest velocity RMSE, surpassing MLP-C even under nominal evaluation. This suggests that exposure to actuator degradation during training enables the node-edge-face policy to learn a broadly effective locomotion strategy rather than one specialized only to damaged conditions.

\section{Conclusion}

We investigated whether higher-order morphology can serve as an inductive bias for quadruped locomotion under compound actuator degradation. In our experiments, the centralized MLP remains strongest after healthy regime training, whereas degraded regime training shifts the advantage to Hodge-F, indicating that rank-2 limb and body structure matters most when the policy must learn compensatory coordination rather than nominal locomotion.

\bibliographystyle{IEEEtran}
\bibliography{references}

\end{document}